# Dual-Arm Telerobotic Platform for Robotic Hotbox Operations for Nuclear Waste Disposition in EM Sites – 24495

Joong-Ku Lee, Young Soo Park
Argonne National Laboratory

**ABSTRACT**

This paper introduces a dual-arm telerobotic platform designed to efficiently and safely execute hot cell operations for nuclear waste disposition at EM sites. The proposed system consists of a remote robot arm platform and a teleoperator station, both integrated with a software architecture to control the entire system. The dual-arm configuration of the remote platform enhances versatility and task performance in complex and hazardous environments, ensuring precise manipulation and effective handling of nuclear waste materials. The integration of a teleoperator station enables human teleoperator to remotely control the entire system real-time, enhancing decision-making capabilities, situational awareness, and dexterity. The control software plays a crucial role in our system, providing a robust and intuitive interface for the teleoperator. Test operation results demonstrate the system's effectiveness in operating as a remote hotbox for nuclear waste disposition, showcasing its potential applicability in real EM sites.

**INTRODUCTION**

Hot cells and gloveboxes are crucial for handling radioactive and hazardous materials in hermetically sealed, and controlled environments, making them beneficial for various decontamination and decommissioning (D&D) and legacy nuclear waste management sites. However, many waste management sites lack such facilities, relying on labor intensive human operations. Hot cell manipulation and glove box operation require frequent maintenance and have limited manipulation capabilities. DOE-EM sites require enhanced hot cell capabilities to replace humans in hazardous waste material handling, improving safety and task efficiency.

To address these needs, a collaborative research and dev elopement (R&D) project is underway to create a robotized hot cell/glovebox system that can be easily transported to EM sites, performing manipulation tasks for nuclear and hazardous waste handling in the field. Recent advances in robotic systems and digital technologies provide timely opportunity for development. The proposed robotic 'hotbox' includes a robotic manipulator system, transportable work cell structure, and operated under robotic digital twin. This paper focuses on the development of a dual-arm telerobotic system, enabling telepresence and human-like dexterous manipulation capabilities.

This project is a joint effort of a team of ANL, ORNL, University of Illinois – Chicago (UIC), Northwestern University (NWU), and OREM cleanup contractor ORCC. The DOE technology development office (TDO) and OREM, in collaboration with the team, function as an integrated project team. DOE-OREM, as the site owner, has the technical need to limit worker exposure to radioactive contamination and mercury vapor. DOE-EM TDO sponsors the work, overseeing progress and playing a critical role in the project success.

**SYSTEM OVERVIEW**

In Fig. 1, the robotic hot cell system is illustrated, consisting of a portable hot cell structure that integrates a dual-arm telerobotic system and a remote handling system. The remote handling system, primarily actuated by a push-pull chain mechanism, is designed for fundamental tasks such as opening the canister and moving the target to the work platform. Simultaneously, the dual-arm telerobotic system is utilized for more dexterous and complicated manipulation tasks. The development of this dual-arm telerobotic system comprises three key components: 1) remote robot arm platform, 2) teleoperator station, 3) digital twin framework. The remote robot manipulator system is mounted on the mobile base, which is a gantry frame. This mobile gantry frame facilitates the remote robot arm platform to execute manipulation tasks across an extensive workspace. Notably, the entire system operation is also connected based on the robotic digital twin framework.

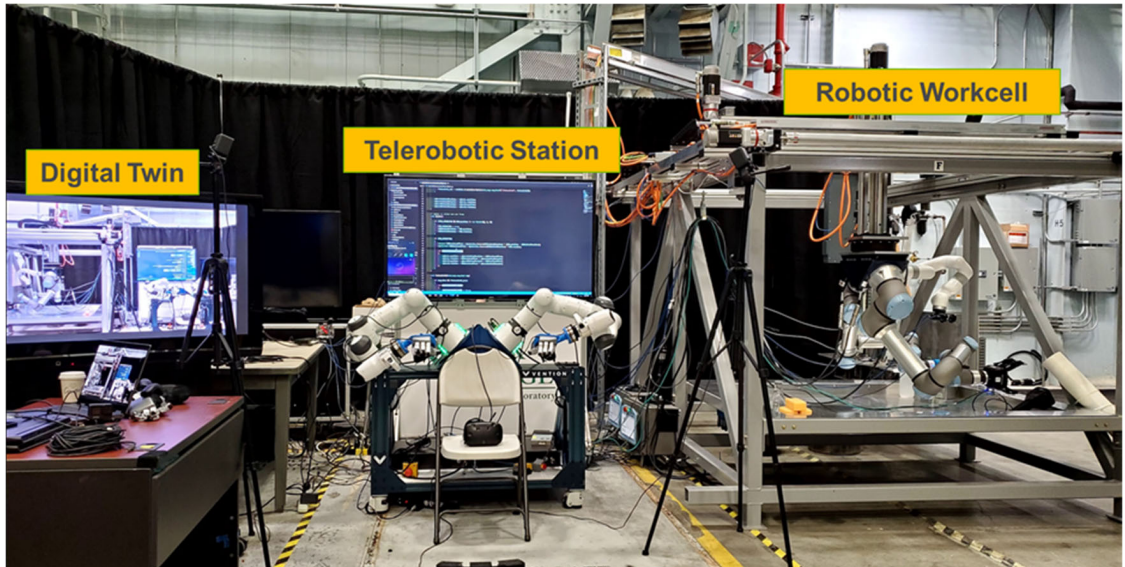

**Figure 1.** Portable robotic hot cell system composed of dual-arm telerobotic system and digital twin.

**Dual-Arm Robotic Platform**

The remote robot arm platform is designed to execute manipulations tasks with a level of dexterity and versatility comparable to a human. This platform includes three primary components:1) dual-arm robot system, 2) dexterous robotic hand, and 3) camera system, as illustrated in Fig. 2.

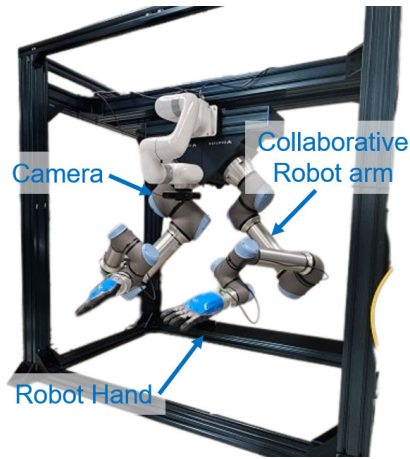

**Figure 2.** Proposed remote robot arm platform for robotic hot cell system. Robot arm, robot hand, camera, and mobile frame are integrated.

*Dual-Arm Robot System*: A recent advancement in robotic arm technology has brought collaborative robots into the commercial market, providing joint torque sensing and precise control capabilities. Leveraging this collaborative robot technology, we implemented an advanced robot control approach to create a dual-arm robot system capable of performing human-like bimanual operations. Specifically, for this project, we opted for Universal Robot's UR5e for the left arm and UR16e for the right arm. It's noteworthy that any collaborative robot can be utilized in a dual-arm system, contingent on the specific task requirements within the hot cell operation.

*Dexterous Robotic Hand*: The dual-arm robot system is equipped with a human-like multi-fingered robotic hand, enabling robots to perform dexterous manipulation akin to humans. In this project, we conducted tests with two types of robotic hands. The first is the qbSofthand2 Research robot hand, capable of simple grasping and opening motions, featuring indirect force feedback through motor current. The second is the Shadow Dexterous Hand, capable of fully actuating each finger joint and providing direct force and tactile feedback through BioTac sensors [1]. The integration of these robotic hands enhances the versatility of our robot arm platform, allowing for a wide range of manipulation tasks.

*Camera System*: Effective visualization of the remote scene is crucial for teleoperation. Within our hot cell structure, multiple camera system is deployed. Some cameras offer a panoramic view of the entire operational scene, while others are dedicated to facilitating remote handling. A pivotal component of dual-arm telerobotic system is the head camera, designed to mimic the function of the human eye during dual-arm operations. The head camera comprises two stereo cameras mounted on a head robot arm that tracks the teleoperator's head pose. Consequently, the head camera, affixed to the head robot arm, follows the teleoperator's head movements, providing a real-time visual representation of the remote scene captured by the head camera.

**Teleoperator Station**

To facilitate remote control of the dual robot arm platform from a distance, we developed a teleoperator station. This station empowers the teleoperator to provide multi-modal perception-action. There are three main components in this system: 1) Local haptic device, 2) Haptic glove, and 3) Head-mounted VR display.

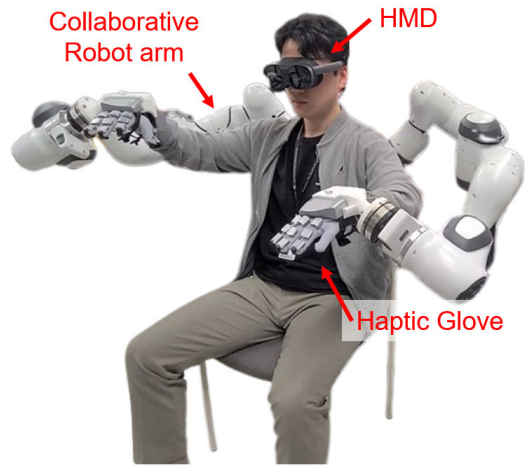

**Figure 3.** Teleoperator station for the teleoperation of robotic hot cell system. Teleoperator interacts with teleoperator station to provide the movement to the dual-arm telerobotic system.

*Local Haptic Device*: A local device capturing the movement of the teleoperator's arm is crucial for the teleoperation system. Additionally, to achieve realistic and transparent teleoperation, a local device capable of rendering force feedback is indispensable. In our setup, we've implemented a bimanual local haptic device utilizing two commercial collaborative robot arms, specifically the Franka Emika Research3. The configuration of the robot arms has been optimized to cover the full range of human arm motion and to provide sufficient haptic feedback force/torque range for the teleoperator [2]. Additionally, our local haptic device is equipped with an FT sensor in its wrist, enabling the implementation of advanced control algorithms that enhance the ease of movement for the human teleoperator.

*Haptic Glove*: To command the dexterous robotic hand within the remote robot arm platform, we've integrated a haptic glove with the local haptic device to capture the movement of the teleoperator hand. Additionally, the haptic feedback obtained from the dexterous robotic hand on the remote side is transmitted to the haptic glove. This integration allows the teleoperator to perceive and feel the forces experienced by the remote robot hand, enhancing the teleoperator's realistic and transparent interaction with the remote environment.

*Head-Mounted VR Display*: The real-time stereo images captured by the head camera system are presented through a head-mounted VR display. Simultaneously, the pose of the teleoperator's head is tracked by a head-mounted device and transmitted to the head robot arm. This enables the remote-side head robot arm to mirror the movement of the teleoperator's head, creating a synchronized and immersive visual experience for the teleoperator during the teleoperation process. In order to reduce the effect of latency during the teleoperation scenario, spherical rendering algorithm [3] to visualize the remote scene is implemented.

**SOFTWARE ARCHITECTURE**

The core of our dual-arm bilateral telerobotic system lies in the software, as we intentionally opted for off-the-shelf hardware in designing the entire system for reproducibility and sustainability. Our key software components include the robot arm control and robot hand control. It should be noted that Robot Operation System (ROS) [4] is utilized as a middleware to communicate each function blocks.

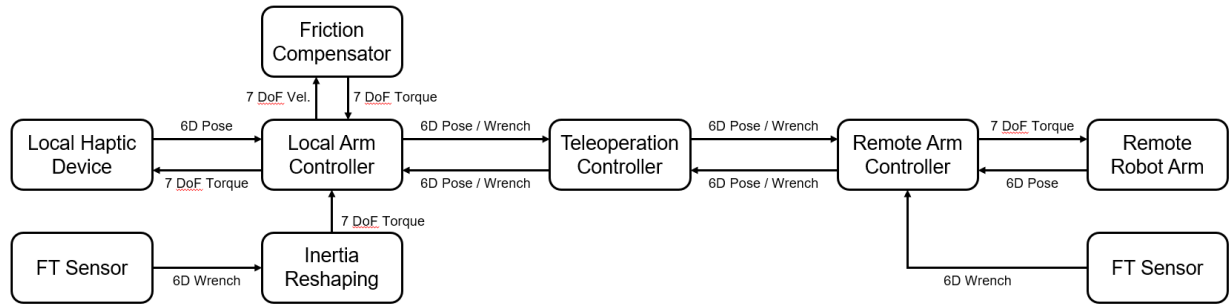

**Figure 4.** The overall block diagram of our robot arm teleoperation controller.

**Robot Arm Control**: In our dual-arm bilateral telerobotic system, the control of the robot arm is governed by three primary controllers. Firstly, the local haptic device robot arm controller executes to directly engage with the teleoperator. Advanced robot control algorithms have been implemented to enhance the teleoperator's movement, providing a sense of ease and lightness during interactions with the local haptic device. In particular, inertia reshaping control has been applied, utilizing force/torque sensors to reduce the teleoperator's perception of robot inertia [5]. Additionally, we've integrated interaction energy-based friction compensation to reduce the friction on the robot joints effectively [6]. These enhancements contribute to a more intuitive and responsive interaction to the local haptic device.
Subsequently, the robot arm controller for the remote robot employs an admittance controller [7] to introduce a level of compliance. The equation for our admittance-control algorithm can be written as

$$M\ddot{x}_d + B\dot{x} + Kx = f_{ext}$$

Where $(x, \dot{x})$ is the current state of the robot end-effector and $f_{ext}$ is the force applied by the user measured by the wrist ft sensor. Solving this equation, we get

$$\ddot{x}_d = M^{-1}(f_{ext} - B\dot{x} - Kx).$$

Using the resulting $\ddot{x}_d$ as a desired end-effector acceleration, we could achieve compliance of the end-effector responding to the external force as mechanical system composed of $M, B, K$. This is crucial, considering our anticipated manipulation tasks may involve interactions with the remote environment.

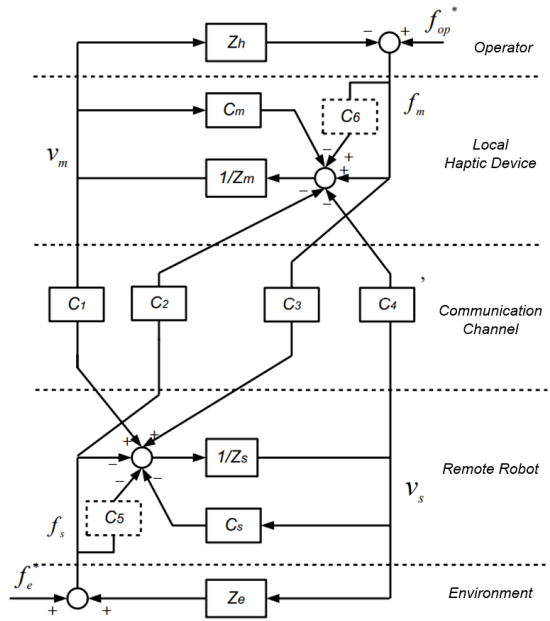

**Figure 5.** Common bilateral teleoperation controller architecture. The 4-Channel bilateral teleoperation architecture is employed when all four control channels $(C_1, C_2, C_3, C_4)$ are present. A position-force architecture is implemented with the existence of $C_1$ and $C_2$.

Finally, the bilateral teleoperation controller, positioned between the remote robot controller and the local haptic device controller, acquires information from both sides and generates control inputs for each side. There are various architectures for bilateral teleoperation. In this project, we have implemented both position-force and 4-channel architecture [8, 9]. As shown in Fig. 5, the four-channel bilateral teleoperation architecture is employed when all four control channels $(C_1, C_2, C_3, C_4)$ are present, while position-force architecture is employed when velocity feedforward channel and force feedback channel $(C_1, C_2)$ are present. The position-force architecture offers the advantage of a lighter feel for the operator, while the 4-channel architecture excels in stability during contact with the environment. In our system, the teleoperator can manually choose between these two modes during teleoperation, providing flexibility to adapt to different task requirements.

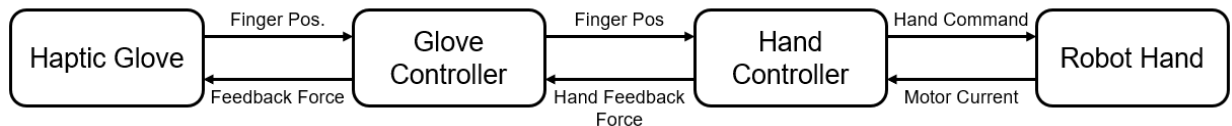

**Figure 6.** The block diagram of our robot hand teleoperation controller.

**Robot Hand Control**: The bilateral control between the haptic glove and the robot hand is implemented in our dual-arm bilateral telerobotic system. In this setup, both the haptic glove and robot hand are controlled by its own driver, with a position-force bilateral control architecture facilitating information exchange between them. The haptic glove conveys finger position information to the remote hand controller, which then outputs the desired finger position for the remote robot hand. Simultaneously, the remote robot hand provides force feedback, estimated using motor current, to the haptic glove controller. The glove controller then outputs the rendered force feedback to the haptic glove, completing the bilateral control loop.

**TEST OPERATION**

All component system developments, encompassing robot arms, hands, head camera, gantry remote systems, have been integrated into a robotic work cell platform (Fig. 7) and tested for functional operations in a non-radiation environment. The demonstration involved a scenario of surrogate waste handling process, which involved remote handling process and robotic manipulation process in the hot cell.

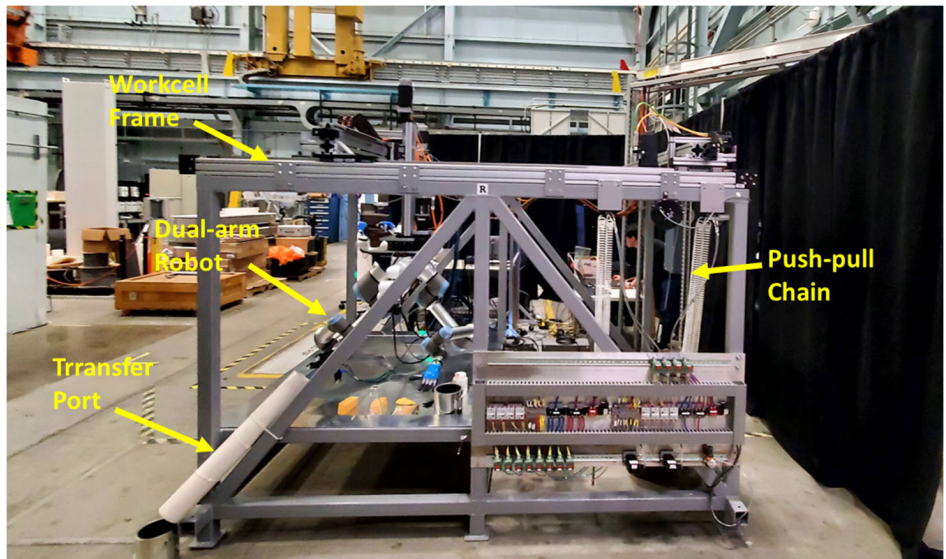

**Figure 7.** Integrated Robotic Work Cell Platform

The remote handling operations use a push-pull chain mechanism with various remotely exchangeable grippers to perform various remote handling of the waste canisters (Fig. 8). The demonstration included opening the canister cover, taking smear samples, and removing the contents from the canister. The smear samples and waste contents are moved to the robotic hot cell floor and collected into a plastic bottle and a waste transfer can.

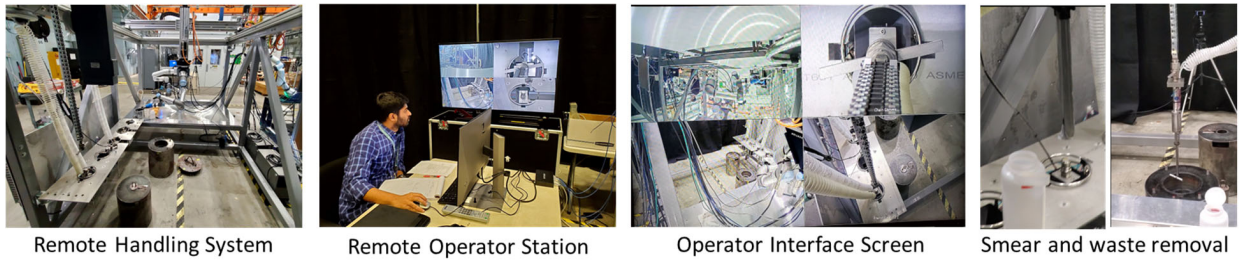

**Figure 8.** Remote Handling System Operation

Subsequently, the dual-arm robotic system was used carry out dexterous operations for waste packaging and disposition in the contained hot cell area. The test operation was performed in telerobotic operation mode as illustrated in Fig. 9. The demonstrated robotic operation includes the following tasks.

Task 1: Grab and move the smear sample bottle to the robot's work area.
Task 2: Close the lid of the sample bottle.

Task 3: Take radiation measurement with a radiation detector.
Task 4: Transfer the sample bottle to the transport port.
Task 5: Grab and move the waste can to the robot's work area.
Task 6: Take a radiation measurement of the waste can.
Task 7: Move the waste can to the put away area.
Task 8: Handling liquid waste
Task 9: Waste collection, sorting, packaging

Fig. 9 shows screenshot of an operator during the test operation.

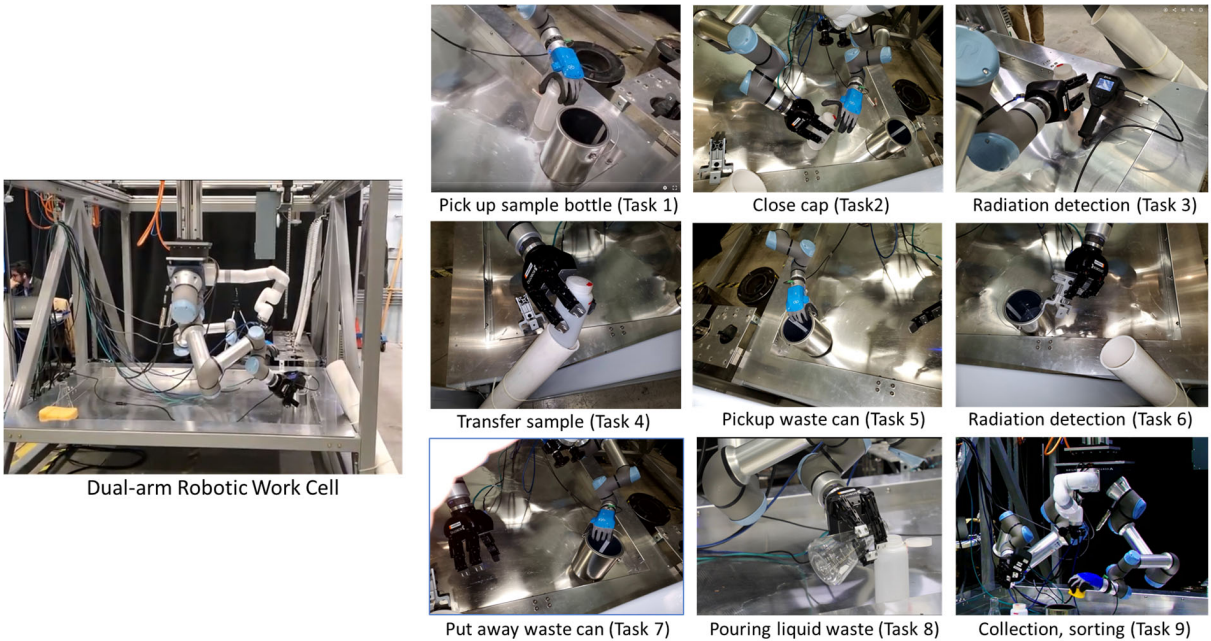

**Figure 9.** Demonstration of Dexterous Robotic Operation for Waste Disposition

**Test Results**:

During the test operations, all operational data, including camera, robot hand motion, and robot hand contact force, have been recorded. It was possible to successfully perform the full sequence of scenario operations. The key attributes of telerobotic performance were due to feasible force reflection, and dual-arm collaborative operations. Fig. 10 shows some test results for some tasks.

Task 1 involved picking up a sample bottle and moving to another location. As illustrated in Fig. 10(a), this task was carried out with one arm, and involved free motion and contact manipulation. During contact manipulation, i.e. picking up the bottle, increase in contact force was identified due to the contact against the work cell floor. This force was reflected to the operator's haptic feedback via bilateral control, and it was possible to adjust control parameters to maintain motion stability.

Task 2 involved closing the bottle cap, while holding the bottle with another hand. This task involved two-arm collaboration, as well as dexterous manipulation with hands. Contact force was incurred due to the contact interaction with the bottle and also between the two arms, which was well handled during the telerobotic operation.

Task 4 addresses putting away the sample bottle into the transfer port. This task involves complex 3-dimensional motion with contact manipulation. Due to the complexity of the required task motion, the initial attempt failed due to unreachable bottle orientation. It was necessary to adjust the bottle grasping configuration with another hand, and it was possible to successfully complete the task. This task performance indicated the resilience of the dual-arm robot system in complex task operation with collaborative motions.

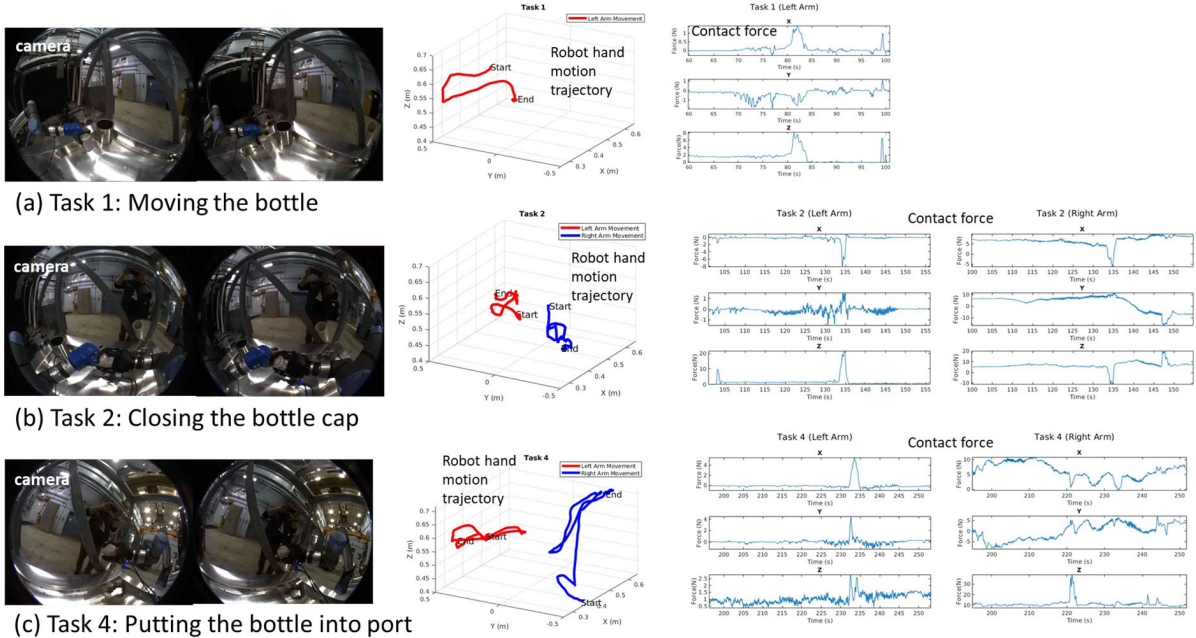

**Figure 10.** Examples of Test Operation Performance

## SUMMARY AND CONCLUSIONS

In this paper, we present the development of a dual-arm telerobotic system, which enables telepresence and human-like dexterous manipulation capability. To achieve effective hot cell manipulation tasks, the development of the proposed telerobotic system includes the innovations on dual-arm robot, teleoperator station design, robot hands, haptic gloves, camera system, and software architecture including control algorithm. We also demonstrated our dual-arm telerobotic system to the test scenario, validating the practical applicability of our system in nuclear disposition missions.

**ACKNOWLEDGEMENTS**

This work is supported by the U.S. Department of Energy, Office of Environment Management, Office of Technology Development. This manuscript has been created by UChicago Argonne, LLC, Operator of Argonne National Laboratory ("Argonne"). The test operation was performed in collaboration with Oak Ridge National Laboratory and United Cleanup Oak Ridge. Argonne is a U.S. Department of Energy Office of Science laboratory.